# LARGE LANGUAGE MODELS "AD REFER-ENDUM": HOW GOOD ARE THEY AT MACHINE TRANSLATION IN THE LEGAL DOMAIN?

VICENT BRIVA-IGLESIAS

vicent.brivaiglesias2@mail.dcu.ie
Dublin City University

JOÃO LUCAS CAVALHEIRO CAMARGO

joo.cavalheirocamargo2@mail.dcu.ie
Dublin City University

GOKHAN DOGRU

gokhan.dogru@uab.cat
Universitat Autònoma de Barcelona

**Abstract**

This study evaluates the machine translation (MT) quality of two state-of-the-art large language models (LLMs) against a traditional neural machine translation (NMT) system across four language pairs in the legal domain. It combines automatic evaluation metrics (AEMs) and human evaluation (HE) by professional translators to assess translation ranking, fluency and adequacy. The results indicate that while Google Translate generally outperforms LLMs in AEMs, human evaluators rate LLMs, especially GPT-4, comparably or slightly better in terms of producing contextually adequate and fluent translations. This discrepancy suggests LLMs' potential in handling specialized legal terminology and context, highlighting the importance of human evaluation methods in assessing MT quality. The study underscores the evolving capabilities of LLMs in specialized domains and calls for reevaluation of traditional AEMs to better capture the nuances of LLM-generated translations.

**Resumen**

Este estudio evalúa la calidad de la traducción automática (TA) de dos grandes modelos de lengua de última generación frente a un sistema tradicional de traducción automática neural (TAN) en cuatro pares de idiomas en el ámbito jurídico. Combinamos métricas de evaluación automática con una evaluación humana de traductores profesionales mediante el análisis de la clasificación, la fluidez y la adecuación de las traducciones. Los resultados indican que, mientras que Google Translate suele superar a los grandes modelos de lengua en las métricas automáticas, los evaluadores humanos valoran a los grandes modelos de lengua, especialmente a GPT-4, de forma comparable o ligeramente mejor en cuanto a fluidez y adecuación. Esta discrepancia sugiere el potencial de los grandes modelos de lengua para trabajar terminología jurídica especializada y contextualizada, lo que pone de relieve la importancia de los métodos de evaluación humana a la hora de evaluar la calidad de la TA. El estudio subraya la evolución de las capacidades de los grandes modelos de lengua en dominios especializados y aboga por una reevaluación de las métricas automáticas tradicionales para captar mejor los matices de las traducciones generadas por grandes modelos de lengua.

## 1. Introduction

Large language models (LLMs), an advanced and complex artificial intelligence (AI) application built upon a vast amount of data for the generation of text and images, which can be particularly useful for language-related tasks (Naveed *et al.* 2023), have been the focus of attention in recent AI progress/advancement in both industry and academia. Since the appearance of BERT (Devlin *et al.* 2019), the application and potential of these LLMs, created from billions of data parameters in computationally intensive training processes, have multiplied and extended to multiple domains, such as healthcare (Kung *et al.* 2023) or the legal field (Trautmann, Petrova & Schilder 2022). The capabilities of these models have been increasing with the emergence of language models trained with larger amounts of data, both for written text (Brown *et al.* 2020) and for spoken text or audio (Radford *et al.* 2022). However, the biggest disruption has been caused by the release of ChatGPT[1], which sparked the interest of the general public in these tools, and extended their use beyond research and industry, bringing them closer to everyday use by laypeople, such as suggested and tested by Yue *et al.* (2023).

With the current hype about the capabilities of these tools and all the attention of media, academia and industry focused on AI and LLMs, studies have recently appeared on their application and disruption in almost all areas of our lives, from teaching (Kasneci *et al.* 2023) to programming (White *et al.* 2023), or many other fields of work. However, there are also many voices that have raised the alarm about the potential dangers of these new technologies in the workplace and the potential loss and/or disruption of jobs due to automation (Eloundou *et al.* 2023), the safety risks of following AI-generated advice and recommendations (Oviedo-Trespalacios *et al.* 2023), as well as the ethical (Zhuo *et al.* 2023) and privacy (Sebastian 2023) risks that may arise in the not too distant future, calling for greater regulation and control of these technologies (Hacker Engel & Mauer 2023).

Given the current context of AI research focusing on LLMs and their use in different scenarios, this article aims to analyse the capacity of this new technology in the field of multilingual specialised communication and, in particular, with respect to the quality of machine translation (MT) in the legal domain across four language pairs. Although LLMs have been receiving a lot of attention,

---

[1] Release blog post of ChatGPT by OpenAI. Online: https:/openai.com/blog/chatgpt (last accessed: 07/06/2023).

AI has already been in use in translation through neural machine translation (NMT) for quite some time, and its use and assessment can be implemented in the same way (Ragni & Vieira 2022: 148). Thus, a quality evaluation of the capabilities of LLMs enables us to ascertain whether they perform better or worse than the current AI technologies in use. Will LLMs with their decoder-only structure (Hendy *et al.* 2023) be better machine translators in the legal domain than traditional NMT systems with an encoder-decoder architecture?

Thus, we first present a literature review of this new technology, which is in its infancy, and its MT capabilities. We then analyse the potential disruption of AI and LLMs and its potential application in the legal domain. We follow with a comparison of translation quality of several state-of-the-art MT systems: a proprietary LLM (GPT-4), an open-source LLM (VICUNA[2]) and a traditional encoder-decoder NMT system (Google Translate[3]). First, results from four automatic evaluation metrics (AEMs) are described to gauge the relative strengths and weaknesses of each system. Second, we share the results of a human evaluation (HE) conducted by translators with professional experience in Spanish, Catalan, Turkish and Brazilian Portuguese. We finally conclude with a discussion of the results and the implications of using LLMs as the MT system for multilingual communication in the legal domain.

## 2. Related work: The disruption of AI and MT in the legal domain

The development of MT includes quality assessment as a crucial aspect that both academia and industry work on (Way 2020), becoming its own subfield in MT research (Castilho & Caseli 2023). Evaluation can be performed through HE and AEMs, with varied practices for different contexts (Castilho *et al.* 2018). The improvement in quality of MT systems in the legal field and their adoption in multiple fields, whether in general industry (ELIS 2022), patent institutions like the World International Property

---

[2] Product page of VICUNA. Online: https://lmsys.org/blog/2023-03-30-vicuna/ (last accessed: 07/06/2023).
[3] Google Translate. Online: https://translate.google.com/ (last accessed: 07/06/2023).

Organization[4], or international institutions such as the European Commission and the creation of eTranslation[5], a public MT system for the legal field, have also led to the analysis of the use of these MT systems in legal institutions (Cadwell *et al.* 2016; Lesznyák 2019; Rossi & Chevrot 2019).

In terms of literature, MT in the legal world has been observed from different points of view. Firstly, focusing on the quality of automatic systems, such as Killman (2014) and the use of MT in Spanish Supreme Court judgments. Another example is that of Wiesmann (2019), who analysed how NMT worked for translating Italian legal texts into German. In addition, Mileto (2019) worked with students to explore their opinions on the use of MT in legal translation, or Briva-Iglesias (2021), who also compared the translation quality of MT engines in the legal field with the translation quality of students specialising in legal translation and refers to the application of MT in the legal language and the constant pressures from the language services industry towards a shorter turnaround model. The other point of view has been computational, and different attempts have been made to improve MT from the technical side to overcome some of the complexities characteristic of legal language (Gotti *et al.* 2008; Koehn & Knowles 2017) or the comparison of MT systems with high- or low-resourced language combinations (Bago *et al.* 2022; Sosoni, O'Shea & Stasimioti 2022).

As much as MT has been adopted, the latest paradigm in MT, neural systems, however, produces different types of errors, such as grammatical errors (Koehn & Knowles 2017) and semantically inadequate words (Raunak, Menezes & Junczys-Dowmunt 2021). That potentially leads to risks in certain scenarios, such as the legal field (Vieira, O'Hagan & O'Sullivan 2021), which complicates the already challenging phenomena of anisomorphism, partial or zero equivalence, or differences between different legal systems during legal translations (Sarcevic 1997; Engberg 2020). Further, the practice of post-editing (O'Brien 2022) has been occurring in the field as a standard practice or to speed the process (Vardaro *et al.* 2019; Killman & Rodríguez-Castro 2022).

---

[4] WIPO Translate, and MT system used for patent translations. Online: https://www.wipo.int/wipo-translate/en/ (last accessed: 07/06/2023).
[5] The MT system developed by the European Comission, eTranslation. Online: https://commission.europa.eu/resources-partners/etranslation_en (last accessed: 07/06/2023).

This indicates that the legal field has made use of AI for language-related tasks focused on translation, but other language tasks have arisen interest as well, such as AI tools providing predictions of judgements by presenting them with a specific and detailed legal case (Long *et al.* 2018), to facilitate the understanding of large amounts of documents in an unknown language and screen important information that needs to be translated by a human (this is also called "e-Discovery"; *cf.* Grossman & Cormack 2010), or smart contracts, which have also gained momentum and relevance recently (Clack 2018). However, the most recent and complex application, LLMs, can perform a series of text generation tasks, which include tasks similar to the ones mentioned previously (Naveed *et al.* 2023), including translation. Because LLMs may also be used for translation (Jiao *et al.* 2023), studies have begun to compare LLMs with NMT systems, both at sentence and document-level (Castilho *et al.* 2023; Wang *et al.* 2023; Zhang, Haddow & Birch 2023).

LLMs are characterised by the implementation of a machine learning technique known as few-shot learning (Brown *et al.* 2020). This approach enables AI applications to perform tasks proficiently with a minimal number of training examples, utilising specific instructions or commands referred to as prompts. Unlike NMT systems, which necessitate extensive training and fine-tuning on substantial datasets and often show limited flexibility beyond their trained purpose, LLMs demonstrate greater adaptability. LLMs are trained on an even greater and diverse linguistic corpora in comparison to NMT systems, endowing them with a more comprehensive understanding of language and context. While NMT systems may be highly specialised and optimised for specific translation tasks, LLMs offer a broader range of applicability due to their ability to quickly adapt to new tasks with minimal additional training while also accounting for context in the tasks. Thus, LLMs may carry tasks with one-shot (one example) or few-shot (some examples), decreasing the need for task-specific data. For example, Han *et al.* (2021) demonstrate such a technique to produce unsupervised MT output through GPT3 and prompting, which essentially is MT without parallel corpora, usually a requirement of NMT systems. Furthermore, the literature demonstrates potential in LLMs for translation, such as i) stylized MT, where style, genre, register, or dialect may be customised for the output through prompting (Lyu, Xu & Wang 2023); ii) translation memory-based MT with LLMs (*Ibid.*), which has potential to improve the perfor-mance of the translation with fuzzy matches (Moslem *et al.* 2023); or iii) hybrid approaches combining the strengths of LLMs with NMT (Hendy *et al.* 2023).

Although quality evaluation has remained the same with the use of NMT (Ragni & Vieira 2022), we have yet to examine if that can be performed in the same way with LLMs, as there is a potential issue of "existing evaluation metrics may not be sufficient to capture the full range of translation quality" (Lyu *et al.* 2023: 4). For this reason, different works approaching translation with LLMs can provide insights in carrying out the evaluation. Hendy *et al.* (2023) conducted sentence-level and document-level evaluation. In the case of the former, different AEMs were used, namely COMET-22 (Rei, C. de Souza *et al.* 2022), COMETkiwi (Rei, Treviso *et al.* 2022), SacreBLEU (Post 2018) and ChrF (Popović 2015), while in that of the latter, an adaptation of COMET was used. Although HE remains the gold standard or the norm for obtaining reliable results (Läubli *et al.* 2020), the evaluations by Hendy *et al.* (2023) were exclusively automatic. Wang *et al.* (2023) also evaluated LLMs using sentence-level and document-level AEMs, specifically, sacreBLEU (Post 2018) both at sentence and document-level, TER (Snover *et al.* 2006), and COMET (Rei, Stewart *et al.* 2020). Additionally, they employed specific discourse metrics, such as cTT and aZPT. Focusing on context-related issues arising from LLMs, Castilho *et al.* (2023) employed inter-annotator agreement (IAA) to provide evaluations at the sentence and document levels. The DELA corpus (Castilho *et al.* 2021) also included context-related issues, such as lexical ambiguity and terminology. Finally, for domain-specific tests, Karpinska & Iyyer (2023) tested BLEURT (Sellam, Das & Parikh 2020) and COMET for literary texts and found that LLMs were able to produce translations that were better than those provided by most NMT systems; at least according to these aforementioned AEMs.

Considering the potential use for LLMs and how adaptable they can be for different domains and how the few-shot technique impacts its output, in this article, the legal domain was chosen to test and check its capability. Siu (2023), for instance, focuses on the use of ChatGPT and GPT-4 for professional translators, addressing how LLMs can aid in identifying terms in documents with domain-specific terminology. However, issues can arise if LLMs are used in legal contexts, as evidenced by Noonan (2023), when reporting problems of bias, confidentiality, and privacy because we do not know how data is handled by the systems, which can even generate inaccurate information. However, Noonan (2023) also suggests that LLMs can be useful for training lawyers and law students. Thus, we believe that evaluating the potential of NMT in comparison to LLMs in legal contexts may provide insights in their potential, issues and evaluation approaches.

**3. Methodology**

Inspired by the works mentioned in the previous sections, we aim to compare the quality between two state-of-the-art LLMs and one state-of-the-art NMT system in the legal domain for four language pairs conducting both automatic and human evaluations. While there is potential for using LLMs in different ways for different tasks (e.g., with written or spoken text), this article focuses on using text-to-text LLMs for MT. Therefore, in this section we present the methodology of the study.

*3.1. MT systems analysed*

First, we used GPT-4, the best ranked LLM at the date of writing (May 2023), according to the Chatbot Arena Leaderboard (Zheng *et al.* 2023). This leaderboard analyses the performance of LLMs in different language-related tasks, such as question answering, verbosity and reasoning ability. Being a proprietary model, we do not know the number of parameters, but it has been reported to yield the best results in MT tasks (Jiao *et al.* 2023). We obtained GPT-4 output by creating prompts via the paid API.

Secondly, and to observe if and to what extent there was any difference between proprietary and open-source LLMs, we analysed VICUNA, the open-source LLM with the best average score in the Chatbot Arena Leaderboard on 22 May 2023[6].

Thirdly, and to compare the translation quality of LLMs with a NMT system, we used Google Translate (GT) as our state-of-the-art baseline, since it is one of the most widely used systems worldwide and supports the languages we were interested in: Spanish, Catalan, Turkish and Brazilian Portuguese.

In addition to selecting systems with different characteristics to evaluate the difference in MT quality between a proprietary

---

[6] LMSYS chatbot leaderboard. Online: https://lmsys.org/blog/2023-05-10-leaderboard/ (last accessed: 07/06/2023).

LLM, an open-source LLM and a NMT engine, the selected languages allow us to make an additional analysis according to the volume of data available for each language combination. As a reference, in the OPUS corpora (Tiedemann 2012) on 31 May 2023, the English-Spanish combination had 920.7 M of aligned sentences, and therefore could be considered to be a high-resourced language combination; the English-Brazilian Portuguese combination had 239.9 M of aligned sentences, thus being considered a middle-resourced language combination. Finally, the English-Turkish language combination had 82.8 M of aligned sentences, and the English-Catalan 33.8 M of aligned sentences. Hence, these latter language combinations could be considered as low-resourced. Therefore, the intention was to analyse the MT performance of these different systems with a selection of languages that are resourced to different extents.

*3.2. Text and translation instructions*

The text chosen as a test set was a legal contract in English, for which the length (537 words) and difficulty with a type-token ratio (TTR) of 0.305 were controlled. The TTR is a metric used to measure the complexity of a text, and the lower the TTR, the higher the difficulty of the text. Our TTR indicates that our source text was a highly specialised, complex text from the legal domain. We controlled text complexity and difficulty to analyse MT output under the constraints of legal wording and terminology. The chosen text was entered directly into the Google Translate website and the translation output was generated, while the following prompt was used for both of the LLMs in our study: "Please provide the [Spanish/Catalan/Turkish/Brazilian Portuguese] translation of the following text: [TEXT]", as per the recommendations of Jiao *et al.* (2023). This is recommended to be the best prompt for instructing LLMs to translate. Sharing this information will allow for increasing replicability and reproducibility when comparing texts with similar length and difficulty in the future with newer systems and analyse their changes in quality.

*3.3. Automatic evaluation of translation quality*

We used both human and automatic evaluation methods to compare the performance of each system. Though AEMs are not as reliable as HE methods (Shterionov *et al.* 2018), they yield replica-ble, objective and rapid results that may provide preliminary insights about comparative performance of each system. However, they should be complemented with HE to draw more reliable conclusions about each system. In terms of AEMs, we used 4 metrics to measure the baseline similarity of each system against human-translated reference translations: BLEU (Papineni *et al.* 2002), TER (Snover *et al.* 2006), chrF3 (Popović 2015) and COMET (Rei, Stewart *et al.* 2020). We specifically include COMET, a relatively new AEM, because of its high correlations with human judgements (Kocmi *et al*. 2021).

Thirty segments from the legal domain were translated by professional translators into four target languages. These segments were considered as the gold standard, and were used as reference translations that were then compared to outputs from three different MT systems in Turkish, Brazilian Portuguese, Spanish, and Catalan. We used the graphical user interface of MATEO (Vanroy *et al.* 2023) to upload the study files[7] and calculate the scores per each metric. Aside from calculating the scores for each metric, MATEO allows for taking one particular score of a system as baseline and calculating the significance of its difference against those of other systems. The resulting table illustrates not only the corpus-level results, but also the mean and the 95% confidence intervals that have been calculated with (paired) bootstrap resampling.

*3.4. Human evaluation of translation quality*

Quality evaluation is one of the most discussed and analysed issues in translation and MT research (Castilho & Caseli 2023). HE of quality is expensive, its reproducibility may vary, and is often carried out by non-expert annotators (Castilho *et al.* 2018). As a consequence, evaluation by expert annotators is considered a good practice in the field (Läubli *et al.* 2020) because they may be able to identify errors that students or non-professional annotators might miss, considering non-experts may not have received formal or extended training for evaluation (Doherty 2017).

Thus, we carried out HE of translation quality with 3 professional translators with more than 5 years of professional experience in the language services industry, and higher education degrees in

---

[7] Study files for each language are provided here: https://drive.google.com/drive/folders/1S-JePjC-VhcX4GMTq0kYwi3fVL899hRh?usp=sharing

Translation and Interpreting. One translator performed the Spanish and Catalan evaluation, another one the Brazilian Portuguese evaluation, and a third translator evaluated the Turkish version. TAUS DQF tools (Görög 2014) were used for the HE by following the evaluation methodology of Briva-Iglesias (2022). We run two types of HE, which were the most common according to both industry and academia procedures:

First, we conducted an MT ranking assessment, where the evaluators saw the MT output of the 3 systems being evaluated and had to assign a score from 1 to 3 according to the quality of the different systems. This assessment provided information about which system provided better translations but did not allow for checking to what degree one system was better than the other.

Second, we conducted a second HE in terms of Adequacy and Fluency using a Likert scale of 1-4, where the evaluators had to assign a score in Adequacy and Fluency to each segment. By "Adequacy" we meant the accuracy of a system, i.e., whether the translation respected the message and content of the original text. Adequacy could be assessed with four different scores: "None", "Little", "Most", and "Everything". By "Fluency", on the other hand, we meant whether the system wrote coherent sentences in the target language. Fluency could be assessed with four different scores: "Incomprehensible", "Disfluent", "Good", and "Flawless". To homogenise the annotators' criteria, the annotation guidelines[8] developed by Briva-Iglesias, O'Brien & Cowan (2023) were presented to the evaluators and used.

## 4. Results

Section 4 explains the results of the different translation quality evaluations conducted. First, results from AEMs are presented, which allow us to grasp a fast and general idea of the overall quality of a system. Nevertheless, as commented above, AEMs present some limitations and, therefore, we also present the results of our HE, so we can then compare both types of translation quality evaluation and discuss the global results in Section 5.

---

[8] Guidelines sent to evaluators are available on:
https://zenodo.org/records/7987955, "Translation Quality Evaluation (TQE) guidelines for assessing Adequacy and Fluency".

*4. 1 Automatic Evaluation Results*

Table I below summarises the results of the AEMs for all four target languages to view the results clearly. While a higher score means better performance in COMET, BLEU and chrF2, a lower score indicates better performance in the case of TER score. An asterisk * indicates that a system differs statistically significantly from the baseline ($p<0.05$). The best system is highlighted in bold. When looking at the overall system performance, GT consistently performs well across all languages and metrics, often outperforming the other systems; GPT-4 shows competitive performance but generally falls short of GT's performance, especially in BLEU, chrF2, and TER scores. Vicuna underperforms significantly compared to GT in all languages and metrics. Its performance is also lower than GPT-4 across all comparisons. According to AEMs, fifteen out of sixteen best scores are obtained by GT (BLEU in Brazilian Portuguese, chrF3 in Turkish and COMET in Turkish). The remaining best score is achieved by GPT-4 (TER in Brazilian Portuguese).

| target language | system | comet↑ | BLEU↑ | chrF2↑ | TER↓ |
|---|---|---|---|---|---|
| *Brazilian Portuguese* | Baseline: GT | 85.5 | **34.2** | **57.7** | **53.0** |
| | GPT-4 | **85.7** | 31.2 | 56.7 | 56.9* |
| | VICUNA | 76.4* | 22.9* | 48.5* | 62.6* |
| *Turkish* | Baseline: GT | **89.8** | **26.3** | **59.9** | **64.1** |
| | GPT-4 | 87.2* | 20.7* | 50.9* | 73.3* |
| | VICUNA | 57.2* | 6.5* | 26.8* | 94.9* |
| *Spanish* | Baseline: GT | **84.2** | **28.6** | **57.0** | **53.8** |
| | GPT-4 | 82.3 | 25.8* | 55.7 | 57.5* |
| | VICUNA | 73.4* | 21.3* | 44.4* | 69.4* |
| *Catalan* | Baseline: GT | **84.5** | **24.7** | **54.3** | **59.8** |
| | GPT-4 | 82.6 | 23.0 | 53.8 | 61.8 |
| | VICUNA | 76.3* | 17.8* | 43.8* | 70.1* |

Table 1. Automatic evaluation scores for four language pairs and three MT systems

If we only take AEMs into account, an overview of the performance across languages also highlights the global superiority of

GT. In the case of translation into Brazilian Portuguese, GPT-4 performs slightly better than GT in COMET, but falls behind in BLEU, chrF2, and TER scores, where GT is superior. Vicuna statistically significantly underperforms in all metrics compared to GT, which is the baseline. In the case of Turkish, GT outperforms both GPT-4 and Vicuna across all metrics. GPT-4 and Vicuna statistically significantly underperform compared to GT, with Vicuna showing a particularly large gap. In Spanish translation, GT again leads in all metrics. Again, Vicuna statistically significantly underperforms compared to GT in all metrics. Lastly, in Catalan translation, similar to Spanish, GT outperforms GPT-4 and Vicuna in all metrics. Vicuna underperforms statistically significantly in all metrics compared to GT.

In summary, for our legal test set and according to AEMs, GT appears to be the most robust system across the languages and metrics tested, with GPT-4 being competitive but not consistently surpassing GT. Vicuna shows considerable underperformance in comparison.

*4.2. Human Evaluation Results*

After presenting the results of the AEMs, this section presents the HE scores, so we can triangulate the data and discern which is the MT system that provides the best MT output in the legal domain for the languages analysed. We conducted different types of HE (ranking, adequacy and fluency) for the different target languages (Brazilian Portuguese, Turkish, Spanish, and Catalan). Thus, below you can find a subsection for every type of evaluation. These subsections start with a brief commentary of the global results, followed by an analysis of the results of every language pair.

*4.2.1. Ranking*
In terms of overall MT ranking results, in three of the four target languages analysed (Brazilian Portuguese, Spanish and Catalan), we can observe that the evaluators considered GPT-4 to be the system with the highest number of segments assessed with the ranking 1 score, i.e. GPT-4 offered higher quality segments than GT and Vicuna. Interestingly, these results differ from those reported by the AEMs. For the remaining target language (Turkish), GT was the MT system that received the ranking 1 score more times. However, if we analyse the data further, we can see that in two of the three languages where GPT-4 was ranked as the best

MT system, the difference with the second MT system (GT) was small. These differences are better discussed in the analysis of each target language.

Brazilian Portuguese
Results for Brazilian Portuguese show that GPT-4 and GT are ranked similarly, both obtaining the best ranking in 21 segments out of 30. Nevertheless, GPT-4 has overall a slightly better performance with nine segments obtaining the ranking 2 score, compared to GT, which was only ranked seven times with this score. Vicuna performed the poorest among the three and obtained the worst ranking in 24 out of 30 segments.

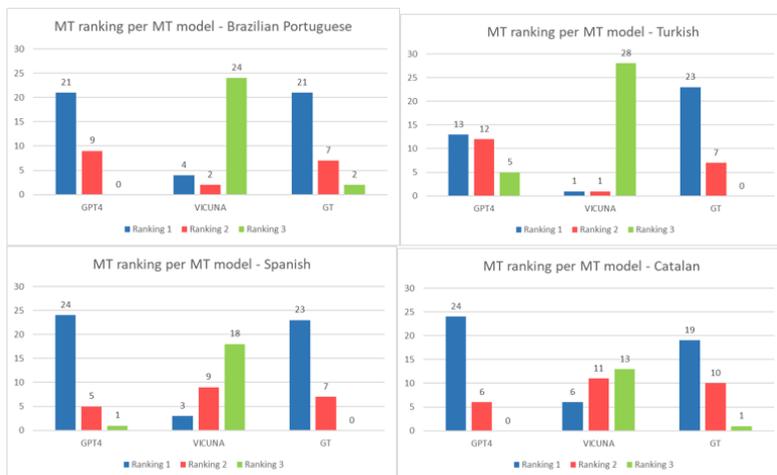

Figure 1. MT ranking per MT system in four language pairs

Turkish
If we analyse the results for Turkish, GT obtained the best ranking in 23 segments out of 30, followed by GPT-4, which obtained the best ranking score 13 times. Vicuna ranked last in nearly all the segments, that is, 28 times out of 30). This finding indicates that, from the perspective of HE, GT outperforms LLM-based MT in this specific language pair.

Spanish
Results for Spanish show that GPT-4 was the system that obtained the most ranking 1 scores (24 out of 30 times), being GT the second MT system with 23 out of 30. This is a close follow-up which indicates that the difference in quality between both systems is not very far from the ranking perspective. As in the previous language pairs analysed, VICUNA lagged behind and only obtained the ranking 1 score three times. In addition, VICUNA was ranked as the worst engine in 18 segments, while GPT-4 only obtained this score once, and GT none.

Catalan
In the English to Catalan language pair, we can observe that GPT-4 distanced itself from GT a bit more than in other language combinations. While GPT-4 obtained the best ranking in 24 segments, GT only did so in 19 segments. It is also worth stressing that GT obtained the worst ranking on one occasion, while GPT-4 did not obtain any ranking 3 score.

*4.2.2. Fluency*

In terms of overall fluency results, GT obtained the best results in Brazilian Portuguese and Turkish, and tied with GPT-4 in Spanish. In Catalan, GPT-4 was the MT system that was assessed with the best fluency scores. Vicuna clearly lagged behind in terms of fluency results in all the target languages analysed. Further discussion of fluency results are included in the analysis of each target language.

Brazilian Portuguese
For Brazilian Portuguese, GPT-4 and GT obtained comparable results in terms of fluency, although the latter generated one segment that was incomprehensible, and GPT-4 only obtained Flawless and Good scores. VICUNA performed the worst, with the least fluent segments in the majority of the segments.

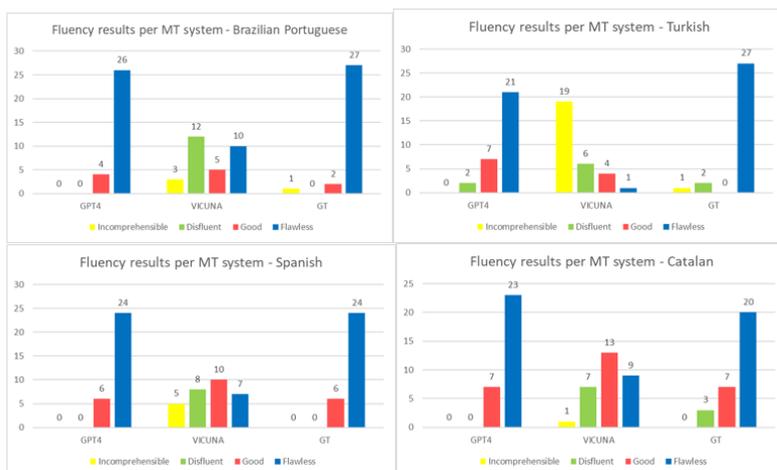

Figure 2. Fluency results per MT system in four language pairs

Turkish
In terms of Turkish, GT obtained the highest fluency score, obtaining a Flawless score in 27 segments, followed by GPT-4, which was assessed as Flawless in 21 segments. Both MT systems had only two Disfluent segments. Vicuna's Turkish translations were mostly incomprehensible, receiving this negative score 19 times.

Spanish
In terms of Spanish fluency results, both GPT-4 and GT obtained identical scores, with 24 segments out of 30 ranked as Flawless and six segments ranked as Good. Once again, VICUNA obtained the worst result, and the Spanish annotator considered that five segments were Incomprehensible.

Catalan
In terms of fluency for Catalan, we can see that GPT-4 was the best ranked system with 23 segments obtaining a Flawless fluency, while the second ranked system, GT, obtained this score in 20 segments. This is not a big difference, but GPT-4 obtained no Disfluent or Incomprehensible scores in any of its segments, while GT obtained three Disfluent scores. Again, VICUNA lagged behind and obtained the worst scores from the different MT systems analysed.

*4.2.3. Adequacy*

In terms of overall adequacy results, GPT-4 obtained the best results in Brazilian Portuguese and Catalan, and tied with GT in Spanish. In Turkish, GT was the MT system that was assessed with the best fluency scores. Vicuna clearly lagged behind in terms of adequacy results in all the target languages analysed. Further discussion of adequacy results is included below.

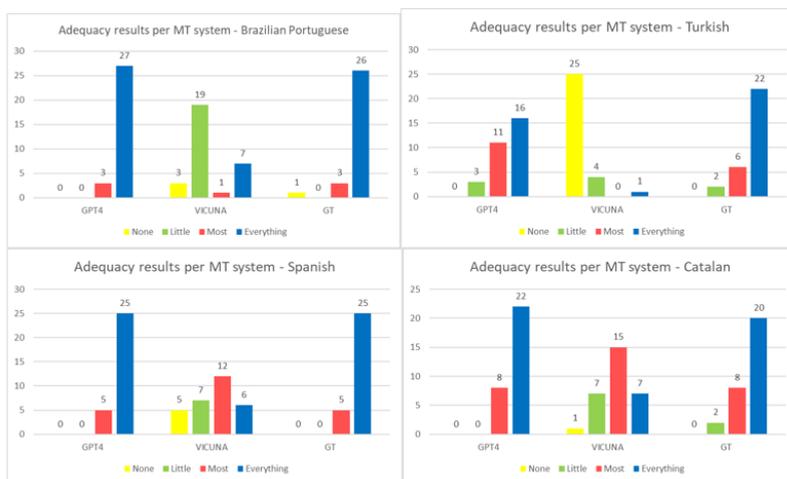

Figure 3. Adequacy results per MT system in four language pairs

Brazilian Portuguese
Adequacy for GPT-4 and GT was also comparable for the English to Brazilian Portuguese language combination, with GT providing only one segment that contained None of the meaning of the source text. This meant that GPT-4 was more consistent than GT for this language combination. VICUNA performed the worst, with most segments scored with Little adequacy.

Turkish
In terms of the adequacy of the Turkish MT output, GT obtained the best scores, with 22 segments assessed as containing all the meaning of the source segment, and six segments respecting most of the meaning. GPT-4 followed, with 16 target segments respecting all the meaning of the source, and 11 target segments having most of the source meaning represented. VICUNA's Turkish translations were marked as totally inadequate in 25 out of 30 segments.

Spanish

In terms of Adequacy results in the English to Spanish combination, GT and GPT-4 obtained once again the same results - with 25 target segments conserving the whole meaning of the source text, and five segments conserving most of the meaning. These two systems obtained no None or Little scores in terms of adequacy, so that meant that the translations produced had high quality. VICUNA, on the other hand, only obtained the best ranking in six of the segments, and the score Most in 12 segments. This meant that 12 of its segments, almost 50% of the total, had little or no adequacy, distancing itself from the other two MT systems analysed.

Catalan
For Catalan, we could observe once again that GPT-4 obtained slightly better results than GT, as the former system had 22 segments ranked with the best score in terms of adequacy, while GT only obtained this score in 20 segments. Nevertheless, it is worth stressing that this difference was not substantial. Again, VICUNA obtained the worst results in terms of adequacy in the English to Catalan language combination.

## 5. Discussion of the results

For Brazilian Portuguese, HE results showed that GPT-4 and GT were consistently good, with GPT-4 providing more consistency in fluency and context than GT, as GPT-4 provided no Incomprehensible or Disfluent segments. On the other hand, GT was assigned one segment with the adequacy score None, and another segment evaluated as Incomprehensible. For the AEMs, GPT-4 scored the best in TER and COMET, while GT scored the best for BLEU and chrF3. This correlates with the HE in the sense that GPT-4 captures more of the context, correlating well with a higher score in COMET. These results demonstrate that GPT-4 might outperform GT in the respect of context for Brazilian Portuguese, as GPT-4 gave no outputs that were incomprehensible or that were unable to carry the meaning into the target. If we look at specific text samples, GPT-4 maintained the terminology translated coherently through the whole document. For instance, "Trust Loan Sellers" was translated as "Vendedores do Empréstimo Fiduciário", while VICUNA could only do so partially as "Vendedores do Empréstimo do Trust". GT, even though it performed better than VICUNA, was not able to translate "Trust Loan Sellers", offering an output identical to the input, not translating it at all.

For Turkish, both HE and AEMs results strongly suggest that GT still outperforms both state-of-the-art LLMs systems: the proprietary GPT-4 and the open-source LLM system Vicuna. However, the scores achieved by GPT-4 were not significantly lower than GT, and when we looked at document-level for terminology, we observed that GPT-4 provided better usage of terms in its adequate context, though it may not be as accurate as GT in sentence level. For example, the term "trust loan" occurred four times in the source text, and it was consistently translated as "güven kredisi" by GPT-4. On the other hand, GT had three different translations for the same source term in the Turkish translation: "güven kredisi", "emanet kredi", "güvenlik kredisi". This capability may be advantageous for LLM systems, since terminology consistency is very important in translation, particularly in legal contexts. Considering that GPT-4 has only been recently launched and only a small percentage of its training data is from Turkish, its translation quality may improve further with new updates and more training data in Turkish and specially in the legal domain.

For Spanish, the HE results indicated that both GPT-4 and GT offered similar quality. Though GPT-4 was in first position in terms of ranking, the difference with GT - the second-best ranked MT system - was only by one segment. Then, if we look at fluency and adequacy results, we can observe that both systems were tied, and obtained the same excellent results. On the other hand, VICUNA obtained poor results, and we can say that it was by far the worst performing MT system in the study. Yet, in terms of AEMs, we can see that the results tended to favour GT.

For Catalan, we can observe that the human annotator ranked GPT-4 to be the best MT system in most of the cases, with a slightly bigger difference than in other language combinations. If we have a closer look at these results from the perspective of adequacy and fluency scores, GPT-4 also maintained its lead with respect to the other systems, though the difference regarding GT was not big. VICUNA lagged behind. While the AEMs demonstrated that GT performed better, GPT-4's scores were not far from those of GT. The performance of the automatic metrics favouring GT mirrored the other language pairs. By examining HE results, the overall better quality is not the only element that is worth stressing from GPT-4, but also global terminology coherence. If we compare contextual terminology consistency throughout the translations, like in

Turkish, we can observe that GT used both "préstec de fideïcomís" and "Préstec Fiduciari" for the source text "Trust Loan". Both options are correct in terms of a legal translation from English into Catalan, but using two terminological choices is inappropriate in some professional translations, as it would be in the case of this contract. On the other hand, GPT-4 consistently translated "Trust Loan" as "Préstec Fiduciari" in every instance this term appeared.

## 6. Conclusion

Our study has been one of the early studies to compare the translation quality of two LLMs against a state-of-the-art NMT system in four different languages to analyse the relative quality of MT capability offered by LLM systems. We chose languages with different characteristics, that is, a high-resourced language like Spanish, a middle-resourced language like Brazilian Portuguese, and two low-resourced languages like Turkish and Catalan. All of these languages were analysed in combination with English. We first conducted an automatic quality evaluation using the most used AEMs in industry and academia, followed by HE with three professional translators, who evaluated the machine translations proposed by the different systems by following the best practices for assessing translation quality (Läubli *et al.* 2020).

By looking at the AEMs alone, we could extract that GT was the best performing system overall in terms of similarity to the reference, gold standard translation. GT obtained the best scores in AEMs in 15 out of 16 evaluations, winning by a landslide. However, it is worth stressing that, while the automatic metrics may provide an initial insight into the performance of the systems, the HE provides a more comprehensive qualitative analysis that investigates further key aspects that a professional translator would check for a legal translation, emphasising terminology or context coherence, for example. As a consequence, HE has been established as the gold standard method for translation quality evaluation by academia and industry.

If we look at the HE results, we can see that GT no longer obtains the best results in the evaluation. In this evaluation task, GPT-4 and GT obtain very similar results. Human evaluators assessed GPT-4 and GT similarly as providing the most accurate and fluent output in most languages combinations analysed (Brazilian Portuguese, Spanish, and Catalan). If there was a difference in these evaluations, it was by a couple of segments, which indicated that the difference was not substantial. Nevertheless, if we looked at the MT output more in-depth, we could see that GPT-4 translated key concepts more consistently throughout the whole document and kept using the appropriate legal terminology. GT, on the other hand, tended to use different terms for the same concept, and was changing its choice throughout the text. Thus, LLMs, GPT-4 in this case, offered better contextual MT capabilities for specialised legal translation.

The only language combination where GPT-4 and GT were not tied was from English into Turkish. HE results suggested that GT was clearly the best performing MT system both in adequacy and fluency. This result may have happened because GPT-4 may contain less Turkish data within its training parameters. As both systems are proprietary, we cannot fully find the actual reason for this result. By following this explanation, it is interesting that Catalan - the other low-resourced language - obtained similar results both for GPT-4 and GT. We think this may be due to the fact that the Catalan language has different open-source communities like Softcatalà[9] or open-source initiatives like AINA[10] that have been making efforts to produce and share high-quality open data for the revitalization of Catalan on the Internet. The fact that Catalan is a Romance language and is more similar to higher resourced languages like Spanish or Portuguese may also be an important factor that helps to improve MT results.

Thus, we can conclude that GPT-4 and GT offered similar MT quality results when translating from English into Spanish, Catalan, and Brazilian Portuguese. The only exception in the languages analysed is when translating into Turkish, where GT clearly obtained better results, and terminology consistency was the only advantage for GPT-4 in this language pair.

Taking these results into account, we suggest using GPT-4 for MT tasks in high-resourced languages (specifically, the above-mentioned languages). In this case, GPT-4 provided users with the advantage of maintaining terminology consistency in specialised domains such as legal texts, possibly due to the LLMs' technique of being a few-shot learner, requiring less training data. This may provide advantages when translating legal texts with new terminology or requirements, which could only be matched by NMT sys-tems trained on specific terminology data or translation memories with large volumes of data. Thus, the flexibility of LLMs provides a strong advantage in specialised domains. Considering this finding, LLMs may be useful to be introduced to acquaint novice translators with terminology consistency, or to train professional translators who are seeking to specialise into legal translation and have little data in this domain either in their available NMT systems or translation memories.

---

[9] Online: https://www.softcatala.org/ (last accessed 07/06/2023).
[10] Online: https://github.com/projecte-aina (last accessed 07/06/2023).

The results of this paper also bring some interesting matters into the scene. If we were to trust AEMs exclusively and blindly, we would clearly see that GT (a traditional encoder-decoder MT system) obtained better results than LLMs. However, after looking at HE results (which have been reported to be the norm and the best practice in translation quality evaluation), we observed that the previous statement was not true. Is there a possibility that AEMs have been developed taking into account how traditional NMT worked and the MT output generated by LLMs escapes to the textual generation form of NMT, therefore being penalised by AEMs? Could this mean that we should re-visit whether LLMs should be assessed with traditional AEMs, at least in the comparison with encoding-decoding systems like NMT? This aspect requires further analysis and study.

In addition, we should comment on the limitations of this piece of research. LLMs are in a period of constant technological development, and a lot has changed from the moment the experiments were first implemented towards the moment of the writing of the results and the review of the paper. As such, the results may not reflect the most current development of the technology. Another limitation is the small test set because we only analysed 30 segments with HE for every MT system per language pair. Larger sample text could have potentially provided more reliability. Yet, this paper is presented as a baseline for reviewing the application of LLMs for performing MT tasks on specialised domains and shares a clear methodology and results that may allow for tracking the development of newer MT technologies in the future.

**BIONOTES**

VICENT BRIVA-IGLESIAS is a researcher on machine translation and human-computer interaction in the Science Foundation Ireland Centre for Research Training in Digitally-Enhanced Reality (d-real), based at the Dublin City University (DCU). In DCU, he lectures the "Localisation" module; at Universitat Oberta de Catalunya, he lectures "Introduction to Python for Translators/Linguists" and "Professional Translation Internships". His research focuses on human-centered AI and machine translation, aiming to augment people's abilities to empower them and reduce their cognitive limitations. His academic experience is influenced by his professional activity, since he runs AWORDZ Language Engineering, a small company that provides language engineering, localisation and internationalisation services.

JOÃO LUCAS CAVALHEIRO CAMARGO has a B. Ed. in Portuguese and English and their respective literatures from Western Paraná State University (UNIOESTE) in Brazil. He holds a Specialist degree in English through distance learning and a Master's in teaching at the same institution. He also holds a Specialist degree in Instructional Design from Instituto de Desenho Instrucional. In his Master's degree research, he designed, implemented and evaluated two translation courses (in-person and distance learning) on translation hermeneutics. He was a Lecturer at the Western Paraná State University, teaching English language teachers, Tourism and Hospitality undergraduates. Currently, he is a PhD student funded by the School of Applied Languages and Intercultural Studies (SALIS) in Dublin City University. His PhD project aims to design, implement and evaluate training on human evaluation of Machine Translation to Master's NLP students.

GOKHAN DOGRU is a visiting postdoctoral researcher at ADAPT-DCU affiliated with the Faculty of Translation and Interpreting at Universitat Autònoma de Barcelona (UAB) in the framework of Margarita Salas Grant. His research interests include terminological quality evaluation in machine translation, different use cases of MT for professional translators and the intersection of translation profession and translation technologies as well as localization.